\documentclass{article}

\usepackage{PRIMEarxiv}

\usepackage[utf8]{inputenc} 
\usepackage[T1]{fontenc}    
\usepackage{hyperref}       
\usepackage{url}            
\usepackage{booktabs}       
\usepackage{amsfonts}       
\usepackage{nicefrac}       
\usepackage{microtype}      
\usepackage{lipsum}
\usepackage{fancyhdr}       
\usepackage{graphicx}       
\graphicspath{{media/}}     
\usepackage[table,xcdraw]{xcolor}
\usepackage{multirow}
\usepackage{caption}
\usepackage{multirow}
\usepackage{booktabs}
\usepackage{tabularx} 
\usepackage{siunitx} 
\usepackage{amsmath}
\usepackage{algorithm}
\usepackage{newfloat}
\usepackage{subcaption}
\usepackage{listings}
\usepackage{algorithmic}
\newcommand{\meanpmstddevbold}[2]{\textbf{#1} $\pm$ \textbf{#2}}      

\title{Deep Clustering via Gradual Community Detection
\thanks{\textit{\underline{Citation}}: 
\textbf{Authors. Title. Pages.... DOI:000000/11111.}} 
}

\author{
  Tianyu Cheng \\
  School of Computer Science\\
  Northwestern Polytechnical University \\
  Xi'an\\
  \texttt{cty18029@163.com} \\
   \And
  Qun Chen \\
  School of Computer Science \\
  Northwestern Polytechnical University \\
  Xi'an\\
  \texttt{chenbenben@nwpu.edu.cn} \\
}

\begin{document}
\maketitle

\begin{abstract}
Deep clustering is an essential task in modern artificial intelligence, aiming to partition a set of data samples into a given number of homogeneous groups (i.e., clusters). Recent studies have proposed increasingly advanced deep neural networks and training strategies for deep clustering, effectively improving performance. 
However, deep clustering generally remains challenging due to the inadequacy of supervision signals. Building upon the existing representation learning backbones, this paper proposes a novel clustering strategy of gradual community detection. It initializes clustering by partitioning samples into many pseudo-communities and then gradually expands clusters by community merging. Compared with the existing clustering strategies, community detection factors in the new perspective of cluster network analysis in the clustering process. The new perspective can effectively leverage global structural characteristics to enhance cluster pseudo-label purity, which is critical to the performance of self-supervision. We have implemented the proposed approach based on the popular backbones and evaluated its efficacy on benchmark image datasets. 
Our extensive experiments have shown that the proposed clustering strategy can effectively improve the SOTA performance. Our ablation study also demonstrates that the new network perspective can effectively improve community pseudo-label purity, resulting in improved self-supervision. 
\end{abstract}


\section{Introduction}
 Due to the success of deep learning, deep clustering has attracted extensive attention from the research community in recent years. The existing approaches for deep clustering can be broadly classified into two categories: two-stage solutions~\cite{dang2021nearest,van2020scan} and single-stage ones ~\cite{caron2018deep,choudhury2019deep,fard2020deep,guo2017improved,li2021contrastive}. The two-stage solutions typically alternate between a feature representation learning stage and a cluster assignment stage, which are usually designed to be separate but complementary. In contrast, the single-stage solutions choose to jointly learn feature representation and cluster assignment within an end-to-end framework. It has been well recognized that both two-stage and single-stage solutions have their own advantages and disadvantages. By separating the stages of representation learning and cluster assignment, the two-stage solutions can easily leverage a wide array of existing techniques on either of them. They also usually require less training cost and can be more easily implemented. In comparison, the single-stage solutions can more effectively align representation learning with clustering objectives, achieving improved performance in some applications. However, it is noteworthy that in both two-stage and single-stage solutions, the efficacy of self-supervision remains as the core challenge of deep clustering~\cite{ren2024deep}. 
 
  To address the challenge of self-supervision, most of the existing studies have focused on designing increasingly advanced representation learning backbones and training strategies. The popular idea was to minimize intra-class distance while maximizing inter-class distance through specific contrastive objectives, e.g., SCAN~\cite{van2020scan} and TCL~\cite{li2022twin}. More recent work proposed to optimize image representation via graph-based contrastive learning, e.g., SDCN~\cite{bo2020structural}, AGCN~\cite{peng2021attention} and CoNR~\cite{yu2024contextually}. Different from basic contrastive clustering that only assumes an image and its augmentation should share similar representations, the graph-based approaches lifted the instance-level consistency to the cluster-level consistency with the assumption that samples in one cluster and their augmentations should all be similar. Some researchers also proposed to exploit text-based supervision signals for deep clustering~\cite{liu2024organizing,li2023image,stephan2024text}. Their general idea was to generate additional semantic information on images via external knowledge bases or pre-trained multimodal large models (e.g., CLIP~\cite{radford2021learning}) and then leverage them to enhance image representation.

  

\begin{figure*}[htbp]
    \centering
    \includegraphics[width=\textwidth]{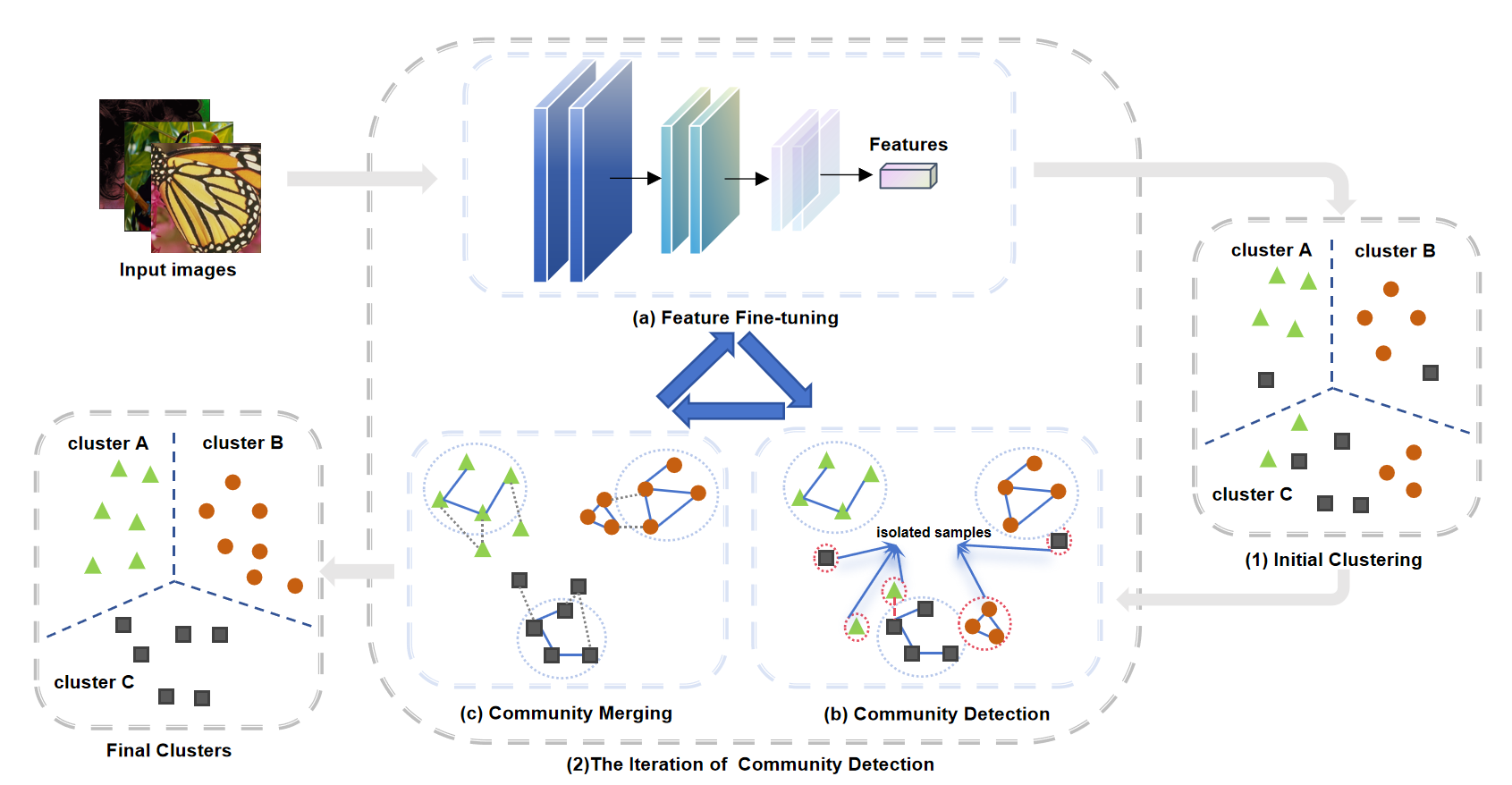}
    \caption{The workflow of the proposed DCvCD: (1) initial clustering; (2) the iteration of community detection: a) fine-tuning an RL backbone using the labeled main communities; b) community detection on unlabeled samples; c) merging isolated communities with main communities.}
    \label{fig1}
\end{figure*}

    Nonetheless, the task of deep clustering generally remains challenging in real scenarios due to the inadequacy of supervision signals. Compared with representation learning, clustering strategy has been much less investigated in the existing literature. In the two-stage solutions, the popular clustering objectives include the traditional K-means loss and maximum mutual information loss~\cite{jiang2017learning}. In the single-stage solutions, clustering strategies are instead implicitly encoded as contrastive losses. These strategies primarily focus on pairwise 
sample relationships and distance-based metrics. Even provided with reasonably good representations, they often group negative samples into a cluster, whose aggregation can adversely affect the performance of self-supervised learning~\cite{misra2020self}.



In this paper, we propose a novel clustering strategy of gradual community detection. Unlike the existing approaches, which directly perform self-supervised learning after initial clustering, our proposed strategy expands clusters by community detection and merging. Community detection brings the new perspective of cluster network analysis into clustering consideration, and can thus improve pseudo-label purity in the self-supervised learning process. We have sketched the proposed approach, denoted by \textbf{DCvCD} (\textbf{D}eep \textbf{C}lustering \textbf{v}ia Gradual \textbf{C}ommunity \textbf{D}etection), in Figure \ref{fig1}. 
After initial clustering, it divides each cluster into many smaller communities by a community detection algorithm. For this purpose, DCvCD employs the Leiden algorithm~\cite{traag2019louvain}, which can theoretically guarantee well-connected internal structures within communities. The Leiden algorithm can effectively overcome the "badly-connected communities" problem often found in traditional detection algorithms, thereby ensuring the robustness and high quality of the final partitioning results~\cite{traag2019louvain}. Then, DCvCD would select the largest community in a cluster as its main community. Finally, it fine-tunes a Representation Learning (RL) backbone by the pseudo-labels in the main communities, and updates feature representation. In the next iteration, it performs a new round of community detection on unlabeled samples, and then merges the resulting isolated communities into the main communities. To improve pseudo-label purity, the algorithm of community merging factors in global structural metrics, e.g., network modularity and average degree, besides the traditional distance metrics. The approach alternates between RL backbone fine-tuning and community merging until all the isolated samples are merged.  


 Built upon the existing RL backbones, DCvCD's main motivation is to improve initial clustering results by community refinement. After initial clustering, the samples within a cluster usually exhibit higher purity compared with before, which makes it easier to generate high-purity main communities. Therefore, in the initialization phase, DCvCD performs community detection on initial clusters instead of an entire dataset. 
  It is noteworthy that DCvCD as a clustering strategy, can in principle work with any RL backbone, including the most recent SOTA ones. Our experiments have shown that by leveraging the recently proposed clustering approaches (e.g., ProPos~\cite{huang2022learning} and CoNR~\cite{yu2024contextually}) for initial clustering, DCvCD can effectively improve the SOTA performance, demonstrating the benefit of leveraging global structural characteristics for cluster assignment. 
  
  It is worthy to point out that the existing work on graph-based contrastive learning primarily focuses on leveraging pairwise node relationships to improve representation learning. Our proposed DCvCD approach instead introduces the new perspective of complex network analysis into clustering consideration and leverages global cluster structural characteristics for cluster refinement. The major contributions of this work can be summarized as follows:

\begin{enumerate}
    \item We introduce a novel clustering strategy of gradual community detection for deep clustering. Incorporating the new perspective of cluster network analysis, it can effectively leverage the concepts of complex network to improve self-supervised clustering. 

   \item We present a novel self-supervised solution of DCvCD for deep clustering, which performs cluster assignment via gradual community expansion. To enhance pseudo-label purity, the algorithm of community merging factors in global structural metrics as well as the traditional distance metrics.  
   
    \item We have implemented the DCvCD solution based on the popular clustering approaches and their RL backbones, and validated its efficacy on benchmark image datasets. Our extensive experiments demonstrate that DCvCD can effectively improve the SOTA performance. Our ablation study has also shown that the introduced perspective of cluster network analysis can effectively improve pseudo-label purity.

\end{enumerate}


\section{Related Work}

Traditional clustering algorithms often perform poorly on complex high-dimensional data, such as images~\cite{cai2009locality,yang2022multi,fang2023joint}. Therefore, deep clustering has established itself as a dominant approach~\cite{ren2024deep}.
The typical roadmap of deep clustering is two-stage. It first leverages deep neural networks to transform complex data into low-dimensional feature representations, and then applies clustering objectives to generate final clusters~\cite{caron2018deep,deng2023strongly,van2020scan,dang2021nearest,niu2022spice}.



A lot of effort has been devoted to designing increasingly advanced DNN backbones for single-stage deep clustering. For instance, Yang et al.~\cite{yang2017towards} proposed the Deep Clustering Network (DCN), which clusters latent features produced by an autoencoder using K-means. Xie et al.~\cite{xie2016unsupervised} used a pretrained autoencoder and iteratively refines clusters by KL-divergence clustering loss assignments. Yang et al.~\cite{yang2016joint} proposed JULE, which combines an agglomerative clustering process with deep learning in a recurrent framework. Guo et al.~\cite{guo2017improved} presented the Improved Deep Embedded Clustering (IDEC) method, which optimizes clustering label assignments and feature representations by local structure of data distribution. Dizaji et al.~\cite{ghasedi2017deep} incorporated cross-entropy loss and cluster-size regularization into deep clustering. Huang et al.~\cite{huang2020deep} proposed a deep clustering method called Partition Confidence Maximization (PICA), which seeks to maximize the global partition confidence of clustering solutions. More recently, Huang et al.~\cite{huang2022learning} proposed the ProPos framework, which balances contrastive and non-contrastive learning through prototype scattering and positive sampling alignment. Most recently, the proposal of GeCC~\cite{chen2025gecc} aimed to enhance clustering robustness of contrastive learning by explicitly modeling domain shifts. In parallel, Chang et al.~\cite{chang2025uncertainty} introduced an uncertainty-aware contrastive framework that models co-cluster probabilities with a Dirichlet distribution to downweight ambiguous pairs, and Wu et al.~\cite{wu2025clusmatch} recast deep clustering as a semi-supervised problem by fine-tuning a pretrained clustering network with unified positive and negative pseudo-labels. Different from these sample-level reweighting and pseudo-labeling schemes, our DCvCD focuses on refining cluster structures by gradually restructuring communities on similarity graphs.

To improve representation learning, some two-stage solutions have adopted an over-clustering strategy. For instance, the Deep Adaptive Clustering method proposed by Chang et al. \cite{chang2017deep} implements a flexible cluster merging mechanism through dynamic adaptive adjustment of features and clusters. Similarly, Huang et al. \cite{huang2020deep} proposed a method of deep semantic clustering by partition confidence maximization for image clustering, ensuring that semantically similar clusters are prioritized for merging.

Orthogonal to the two-stage and single-stage work, the research community has also investigated graph-based deep clustering, which aims to leverage the underlying structure of data for improved performance ~\cite{chiang2019cluster,bo2020structural,peng2021attention}. For instance, Chiang et al. \cite{chiang2019cluster} proposed a fast and memory-efficient method based on Graph Convolutional Networks (GCNs). Bo and Wang \cite{bo2020structural} presented the Structural Deep Clustering Network (SDCN), combining GCNs with the DEC framework to integrate structural information into deep clustering. Peng et al. \cite{peng2021attention} proposed the Attention-driven Graph Clustering Network (AGCN), which dynamically aggregates node attribute features and topological features, and adaptively fuses multi-scale features embedded in different layers. Li et al. \cite{li2022twin} introduced Twin Contrastive Learning (TCL), which simultaneously conducts contrastive learning at both instance and cluster level. More recently, Huang et al. \cite{huang2024deepclue} proposed a deep clustering framework, DeepCluE, which integrates feature representations from multiple network layers, overcoming the limitations of traditional methods that mainly rely on single-level features.
Yu et al. \cite{yu2024contextually} also introduced Contextually Affinitive Neighborhood Refinery (CoNR), which refines cluster assignments by online semantic re-ranking. Moreover, it employed a progressive boundary filtering mechanism to mitigate noisy neighbors.

The above-mentioned works primarily operate in a single-view setting. Orthogonal to this, another emerging research direction is multi-view clustering, where data is represented by multiple feature sets. Recent work in the multi-view domain has tackled challenges such as noisy pseudo-labels (ROLL~\cite{sun2025roll}) and fairness~\cite{xu2025deep}. On the other hand, integrating textual information into image clustering to enhance semantic consistency and interpretability has become a research frontier in recent years. Approaches such as TeDeSC~\cite{liu2024organizing} explore how to discover semantic partitions through text-driven frameworks, while TAC~\cite{li2023image} optimizes clustering decisions using external knowledge (e.g., WordNet). Similarly, Stephan et al.~\cite{stephan2024text1,stephan2024text} employ generated captions and VQA-style descriptions as an intermediate text space for clustering, and Cai et al.~\cite{cai2024enhancing} fuse caption and visual features to obtain more semantically meaningful clusters. In a more controllable setting,  $(IC\|TC)$~\cite{kwon2023image} conditions clustering on user-specified text criteria, and recent cross-modal methods such as SEIC~\cite{li2025self} refine CLIP-based encoders by enforcing consistency between image and text semantics.


\section{Methodology}


 As shown in Figure~\ref{fig1}, in the initialization phase, DCvCD performs community detection within each initial cluster, and divides its samples into multiple pseudo-communities, the largest of which is selected as the main community. In the iterative phase, DCvCD first exploits the pseudo-label information of samples in main communities to fine-tune an RL backbone, and then performs community detection on all the unlabeled samples; finally, it merges pseudo-communities with main communities. Since the DCvCD solution can in principle be built upon any existing RL backbone, and its initial clustering can leverage any clustering algorithm, the rest of this section focuses on the technical solutions of community detection \& merging and backbone fine-tuning. 



\begin{figure}[htbp]
    \centering
    \includegraphics[width=0.4\textwidth]{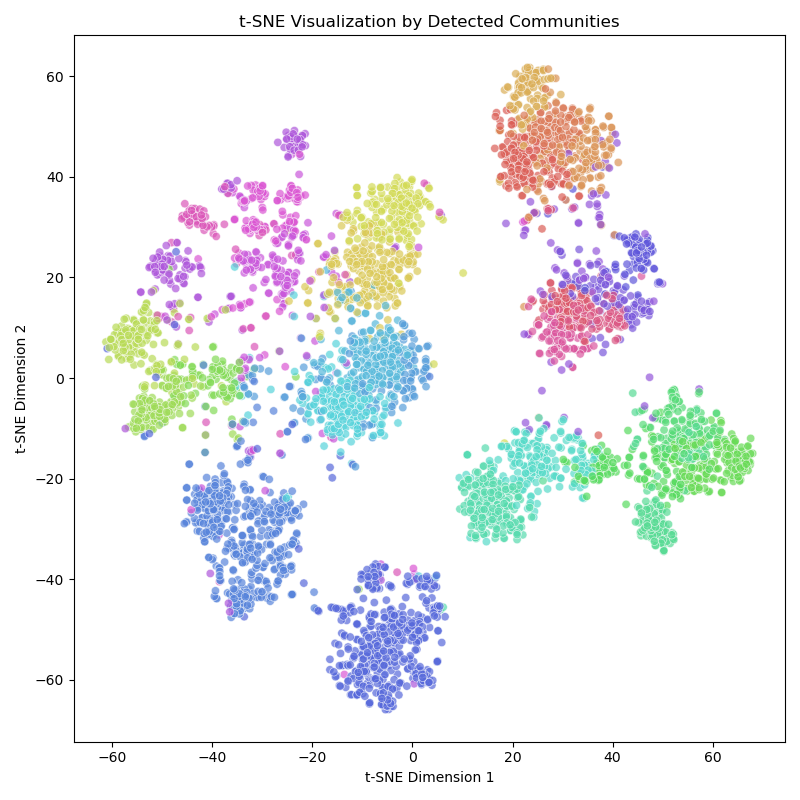}
    \caption{Community visualization after partitioning an initial cluster with the Leiden algorithm: a few major communities and many smaller communities.}
    \label{fig2}
\end{figure}

\subsection{Community Detection \& Merging} 

\noindent\textbf{Community Detection: }
to enable community detection, DCvCD constructs a weighted network of samples within each initial cluster based on their representation similarity, in which edge weights are measured by the cosine similarity between samples. Specifically, it connects two samples by an edge if and only if their representation similarity exceeds a threshold (e.g., 0.5 in our implementation).


Then, DCvCD applies a community detection algorithm on each initial cluster to partition the samples into many pseudo-communities. Our implementation employs the Leiden algorithm. As illustrated in Figure~\ref{fig2}, the core principle of the Leiden algorithm is to ensure that the resulting communities are internally cohesive, typically yielding only a few large communities and a limited number of small ones. DCvCD selects the largest pseudo-community as the initial main community. It is noteworthy that the core advantage of the Leiden algorithm is highly aligned with DCvCD's goal of improving pseudo-label purity through global structure. Increasing modularity can directly enhance the correlation among samples within communities.

Technically, the Leiden algorithm identifies communities by maximizing modularity, which reflects the density of connections within communities versus the sparsity of connections between them. It is noteworthy that, as a more advanced algorithm than the traditional Louvain algorithm, the Leiden introduces a crucial refinement phase in its optimization process. This phase specifically addresses the inherent shortcoming of the Louvain algorithm, which can produce badly-connected communities that are not internally connected. Consequently, the Leiden algorithm theoretically guarantees that all identified communities are internally well-connected. This improvement makes the Leiden algorithm more reliable and robust for the clustering task.

Formally, the objective function optimized by the Leiden algorithm, modularity Q, is defined as:
\begin{equation}
    Q = \frac{1}{2W} \sum_{i,j} \left[ w_{ij} - \frac{k_i k_j}{2W} \right] \delta(x_i, x_j)
    \label{eq:modularity}
\end{equation}
where $W$ represents the sum of all edge weights in the network, i.e., $W = \sum_{i,j} w_{ij}$; $w_{ij}$ is the weight of the edge between nodes $i$ and $j$, with $w_{ij} = 0$ if no edge exists; $k_i$ and $k_j$ are the weighted degrees of nodes $i$ and $j$ respectively, with $k_i = \sum_j w_{ij}$; and $\delta(x_i, x_j)$ denotes the community indicator function, where $\delta(x_i, x_j) = 1$ if nodes $i$ and $j$ belong to the same community, and 0 otherwise.

   To ensure high pseudo-label purity in initial main communities, DCvCD applies risk screening to filter out highly risky samples. Intuitively speaking, a sample is considered as risky if its degree of deviation from the community centroid is beyond a pre-specified confidence interval (e.g., 0.9 in our implementation) in terms of distance.

\vspace{0.05in}   
\noindent\textbf{Community Merging:} to effectively reduce the accumulation of pseudo-label errors during the iterative process of community merging, we present an ensemble metric, consisting of both network structural indicators and traditional distance indicators, to optimize the merging of isolated communities with main communities.

Intuitively speaking, our solution aims to ensure that a merging operation maintains community structure consistency while minimizing the negative impact of misclassification. Towards this aim, our merging strategy not only considers the representation similarity between communities, but also incorporates network structural metrics such as modularity increment and average degree change to ensure the structural coherence of merged communities. Specifically, we define the guiding ensemble metric by 
\begin{equation}
    L =\frac{\Delta Q}{\max(\Delta Q)} + \frac{\Delta k}{\max(\Delta k)} -\frac{t}{\max(t)}
\end{equation}
where \( L \) denotes the ensemble score, \( \Delta Q \) denotes modularity increment after community merging, \( \Delta k \) denotes average degree change, and \( t \) denotes the measured distance between two communities. We elaborate each measure as follows:
\begin{itemize}
\item \textbf{Modularity Increment \( \Delta Q \).} Modularity increment is used to assess the overall optimization effect of merging on the structure of the main community. The modularity increment, $\Delta Q$ , measures the "quality" improvement of community structure after merging. A higher \( \Delta Q \) indicates that the dense connectivity within a community after merging is significantly higher than expected in a random graph, implying that two communities have strong global similarity.
\item \textbf{Average Degree Change \( \Delta k \).} Besides the global quality improvement as measured by \( \Delta Q \), we also introduce the average degree increment, \( \Delta k \), to measure the change in local connectivity, which can indicate the structural balance of a community after merging. A larger \( \Delta k \) indicates stronger node connectivity between two communities, suggesting higher similarity in their local structures.
\item \textbf{Community Distance \( t \).} As usual, the ensemble metric also uses the node distances between communities to assess positional similarity in the feature space, aiming to avoid the merging of communities that are far apart. Formally, the metric of \( t \) is defined as:
\begin{equation}
    t = \frac{1}{|C_1| \cdot |C_2|} \sum_{x_i \in C_1} \sum_{y_j \in C_2} d(x_i, y_j)
\end{equation}
where $C_1$ and $C_2$ denote two communities, and $d(x_i, y_j)$ denotes the Euclidean distance between two nodes. It can be observed that a smaller \( t \) value indicates that the nodes are closer in feature space, which helps to avoid structural instability caused by long-distance merging. 
\end{itemize}

In each iteration, for each main community, \( C_i\), DCvCD selects an isolated community with the highest \( L \) metric value w.r.t. \( C_i \) and merges it with \( C_i \). After community merging, DCvCD re-trains the RL backbone on the updated main communities, re-constructs a network on the nodes in isolated communities, and re-executes the Leiden algorithm. 



\vspace{0.05in}
\noindent\textbf{Efficiency Analysis:} theoretically, given a network with $n$ nodes and $m$ edges, the average-case time complexity of the Leiden algorithm is approximately $O(m+n\log n)$, making it highly efficient for large-scale networks. Furthermore, it can be observed that after the first iteration, a majority of samples would be assigned to main communities; subsequent iterations would operate on a dramatically reduced number of unlabeled samples. As a result, the overhead of community detection would decrease dramatically after the first iteration. On the other hand, the computational complexity of community merging is approximately $O(n_i \cdot n_j)$, where $n_i$ and $n_j$ denote the numbers of nodes in the two communities respectively. In practice, since the size of most isolated communities is usually very small (i.e. in single digits), its overhead is usually negligible compared with that of community detection. Our efficiency evaluation results presented in the section of empirical study show that most of the DCvCD overhead is spent on backbone fine-tuning, and its total overhead is comparable to that of existing clustering approaches.

\subsection{Backbone Fine-tuning}
During each iteration, since the samples in the main community are regarded as reliable cluster members, DCvCD assigns pseudo-labels to these samples to guide RL backbone fine-tuning. The essential objective of fine-tuning is to make the generated embeddings better reflect the current cluster structure. To this end, as usual, DCvCD uses a clustering-aware loss function. Specifically, it adopts an adapted form of the InfoNCE (Noise-Contrastive Estimation) loss, which can be formally defined as follows:
\small
\begin{equation}
\mathcal{L} = -\sum_{i=1}^{N} \log \frac{\exp(\text{sim}(z_i, z_{p_i}) / \tau)}{\exp(\text{sim}(z_i, z_{p_i}) / \tau) + \sum_{j \in \mathcal{N}_i} \exp(\text{sim}(z_i, z_j) / \tau)}
\end{equation}

\begin{table*}[ht]
\centering
\small
\caption{Comparative results of image-only clustering approaches on five image benchmarks, with the best results highlighted. "---" indicates that the experimental results or source code of these methods cannot be obtained.}
\label{tab1}
\begin{tabular}{lcccccccccccccccc}
\toprule
\multirow{2}{*}{Datasets} & \multicolumn{3}{c}{CIFAR-10(\%)} & \multicolumn{3}{c}{CIFAR-100(\%)} & \multicolumn{3}{c}{STL-10(\%)} & \multicolumn{3}{c}{ImageNet-10(\%)} & \multicolumn{3}{c}{ImageNet-Dogs(\%)} \\
\cmidrule(r){2-4} \cmidrule(r){5-7} \cmidrule(r){8-10} \cmidrule(r){11-13} \cmidrule(r){14-16} 
Metrics & NMI & ACC & ARI & NMI & ACC & ARI & NMI & ACC & ARI & NMI & ACC & ARI & NMI & ACC & ARI\\
\midrule
DCGAN (2015) & 26.5 & 31.5 & 17.6 & 12.0 & 15.1 & 4.50 & 21.0& 29.8 & 13.9 & 22.5 & 34.6 & 15.7 & 12.1 & 17.4 & 7.8 \\
JULE (2016)  & 19.2 & 27.2 & 13.8 & 7.50 & 13.7 & 3.30 & 18.2 & 27.7 & 16.4 & 17.5 & 30.0 & 13.8 & 5.4  &13.8  &2.8
  \\
DCCM (2019)  & 49.6 & 62.3 & 40.8 & 28.5 & 32.7 & 17.3 & 37.6 & 48.2 & 26.2 & 60.8 & 87.1 & 55.5 & 32.1 & 38.3  & 18.2 \\
PICA (2020)  & 59.1 & 69.6 & 51.2 & 31.0 & 33.7 & 17.1 & 61.1 & 71.3 & 53.1 & 60.8 & 90.1 & 82.2 &33.6 & 32.4 &17.9 \\
CC (2021)  & 70.5 & 79.0 & 63.7 & 43.1 & 42.9 & 26.6 & 76.4 & 85.0 & 72.6 & 80.2 & 87.0 & 76.1  &44.5 & 42.9 & 27.4\\
CRLC (2021)  & 67.9 & 79.9 & 66.4 & --- & --- & --- & 72.9 & 81.8 & 68.2 & 83.1 & 85.4 & 75.9 & --- & --- & ---\\
MICE (2021)  & 73.5 & 83.4 & 69.5 & 43.0 & 42.2 & 27.7 & 61.3 & 72.0 & 53.2 & 61.3 & 84.2 & 82.2 &42.3 & 43.9 & 28.6 \\
GCC (2021)  & 76.4 & 85.6 & 72.8 & 47.2 & 47.2 & 30.5 & 68.4 & 78.8 & 63.1 & 84.2 &90.1 & 82.2  &49.0 & 52.6 &36.2 \\
EDESC (2022)  & 46.4 & 62.7 & --- & 37.0 & 38.5 & --- & 68.7 & 74.5 & --- & --- & --- & --- & --- & --- & --- \\
DFVC (2022)  & 64.3 & 75.6 & 61.5 & 43.5 & 47.2 & 26.1 & 64.2 & 73.1 & 59.8 & 75.3 & 85.7 & 73.6 & --- & --- & --- \\
DCSC (2022)  & 70.4 & 79.6 & 64.4 & 45.2 & 46.9 & 29.3 & 79.2 & 86.5 & 74.9 & --- & --- & ---  & --- & --- & ---\\
TCL (2022)  & 81.9 & 88.7 & 78.0 & 50.8 & 51.2 & 36.7  & 79.9 & 86.8 & 75.7 & 87.5 & 89.5 & 83.7 & 62.3 & 64.4 & 51.6\\
SPICE (2022)  &73.4 & 83.8 & 70.5 & --- & --- & --- &81.7 & 90.8 & 81.2 & 82.8 & 92.1 & 83.6 &62.7 & 67.5 & 52.6\\
CCES (2023)  & 72.4 & 81.2 & 69.4 & 44.2 & 43.6 & 30.1 & 77.5 & 84.7 & 73.1 & --- & --- & --- & --- & --- & --- \\
SeCu (2023)  & 79.9 & 88.5 & 78.2 & 51.6 & 51.6& 36.0 & 70.7 & 81.4 & 65.7 & --- & --- & --- & --- & --- & ---\\
DivClust (2023)  & 71.0 & 81.5 & 67.5 & 44.0 & 43.7 & 28.3 & 65.1 & 74.4 & 64.0 & 85.0 & 90.0 & 81.9 &51.6 &52.9 &37.6 \\
ProPos (2023)  & 88.6 & 94.3 & 88.4 & 51.5 & 53.2 & 37.6 & 75.8 & 86.7 & 73.7 & 89.6 & 95.6 & 90.6 &69.2 & 74.5 &62.7 \\
CoNR (2023)  & 86.7 & 93.2 & 86.1 & 51.3 & 53.6 & 37.5 & 85.2 & 92.6 & 84.6 & 91.1 & 96.4 & 92.2 &74.4 & 79.4 & 66.7\\
DeepCluE (2024)  & 72.7 & 76.4 & 64.6 & 47.2 & 45.7 & 28.8 & --- & --- & --- & 88.2 & 92.4 & 85.6 &44.8 & 41.6 & 27.3  \\
GeCC (2025)  & 75.5 &84.4 & 71.0 & 44.1 & 45.1 & 28.7 & --- & --- & --- & 90.9 & 96.1 & 91.6 & 54.9 & 60.3 & 42.1\\
UACL (2025) & 82.5 & 89.9 & 79.4 & 53.6 & 53.9 & 37.3 &  80.2 & 87.1 & 75.8 & 91.3 & 95.3 & 89.6 & 64.0 & 64.6 & 52.9\\
\hline
DCvCD (DivClust) &81.8 & 90.1& 80.2 & 51.3 & 52.2 & 36.8 & 77.6 & 84.1 & 73.9 &89.8 &93.4 &87.4 & 56.9& 59.3 &41.5\\
DCvCD (ProPos) & \textbf{90.7} & \textbf{95.3} &\textbf{89.6} &52.8 & 54.9 & \textbf{39.1} &78.1 &88.9 &75.2 &90.3 &96.0 &90.7 &71.1 & 75.8 &\textbf{68.3} \\
DCvCD (CoNR) &88.7 &94.2 &87.8 &\textbf{53.6} &\textbf{55.8} &39.0 &\textbf{87.2} &\textbf{94.0} &\textbf{85.8} &\textbf{92.0} &\textbf{97.5} & \textbf{93.1} &\textbf{76.2} &\textbf{81.0} & 67.4\\

\hline

\hline
\end{tabular}
\end{table*}
Where \( N \) denotes the number of training samples (i.e., pseudo-labeled samples), \( z_i \) denotes the embedding vector of node $i$, \( z_{p_i} \) denotes the embedding of a positive sample sharing the same cluster with \( z_i \), \( \mathcal{N}_i \) denotes the set of negative samples, \( \text{sim}(\cdot, \cdot) \) denotes a similarity function, e.g. cosine similarity, and \( \tau \) denotes the temperature parameter scaling the similarity.

   After the fine-tuning, DCvCD leverages the updated backbone to refresh the embeddings of the samples in the entire dataset. Then, it invokes a new iteration of community detection, community merging and backbone fine-tuning.

\section{Empirical Study}

   In this section, we empirically evaluate the performance of our proposed approach on benchmark image datasets. 
   
   
   
\subsection{Experimental Setup}

We have conducted experiments using five benchmark image datasets: CIFAR-10 \cite{krizhevsky2009learning}, CIFAR-100 \cite{krizhevsky2009learning}, STL-10 \cite{coates2011analysis}, ImageNet-10 \cite{deng2009imagenet} and ImageNet-Dogs \cite{deng2009imagenet}, whose brief descriptions are as follows: 1) \textbf{CIFAR-10}: containing 60,000 color images in 10 classes, CIFAR-10 is widely used for image clustering evaluation; 2) \textbf{CIFAR-100}: containing 20 superclasses and 100 subclasses, CIFAR-100 is an extension of CIFAR-10. Due to its greater number of categories, CIFAR-100 is more challenging, often used for the evaluation of fine-grained clustering; 3) \textbf{STL-10}: containing 13,000 images across 10 classes, STL-10 is a popular benchmark derived from ImageNet, often used for unsupervised and self-supervised learning; 4) \textbf{ImageNet-10}: as a subset of ImageNet, ImageNet-10 contains 13,000 color images from 10 classes. It is usually used for quick image classification evaluation without requiring full training on the large ImageNet dataset; 5) \textbf{ImageNet-Dogs}: a subset of ImageNet used for fine-grained clustering, this dataset consists of 19,500 images from 15 different breeds of dogs, presenting a significant challenge due to the subtle inter-class variations.

Since traditional approaches, e.g., K-Means~\cite{macqueen1967some}, SC~\cite{zelnik2004self}, AC~\cite{gowda1978agglomerative} and NMF~\cite{zheng2009tumor}, are not competitive compared with deep clustering methods, we compare our proposed DCvCD approach with 21 mainstream deep clustering alternatives, nearly two decades. The compared methods include DCGAN~\cite{radford2015unsupervised}, JULE~\cite{yang2016joint}, DCCM~\cite{wu2019deep}, PICA~\cite{huang2020deep}, CC~\cite{li2021contrastive}, CRLC~\cite{do2021clustering}, MICE~\cite{tsai2020mice}, GCC~\cite{zhong2021graph}, EDESC~\cite{cai2022efficient}, DFVC~\cite{ji2021decoder}, DCSC~\cite{zhang2022improved}, and the most recently proposed SeCu~\cite{qian2023stable}, DivClust~\cite{metaxas2023divclust}, CCES~\cite{yin2023effective}, DeepCluE~\cite{huang2024deepclue}, TCL~\cite{li2022twin}, SPICE~\cite{niu2022spice}, ProPos~\cite{huang2022learning}, CoNR~\cite{yu2024contextually}, GeCC~\cite{chen2025gecc} and UACL~\cite{chang2025uncertainty}. For a supplementary comparison against text-enhanced methods, please see Appendix A. As usual, we measure the performance of deep clustering by three metrics, Clustering Accuracy (ACC), Normalized Mutual Information (NMI) and Adjusted Rand Index (ARI). 
  \begin{table*}[htbp]
    \centering
        \caption{The ablation evaluation results of the clustering strategy w.r.t. various baseline clustering approaches, in which the better results are highlighted}
        \label{tab2}
    \begin{tabular}{lccccc} 
        \toprule
        Method & CIFAR-10(\%) & CIFAR-100(\%) & STL-10(\%) & ImageNet-10(\%) &ImageNet-Dogs(\%)\\
        \midrule
        DivClust & 81.5 & 43.7 & 74.4 & 90.0 &52.9 \\
        DCvCD (DivClust) & \meanpmstddevbold{90.1}{1.2} & \meanpmstddevbold{52.2}{1.8} & \meanpmstddevbold{84.1}{1.0} & \meanpmstddevbold{93.4}{0.4} &\meanpmstddevbold{59.3}{0.7}\\
        \addlinespace 
        CC & 79.0 & 42.9 & 85.0 & 87.0 &42.9\\
        DCvCD (CC) & \meanpmstddevbold{87.8}{0.6} & \meanpmstddevbold{50.5}{0.3} & \meanpmstddevbold{93.1}{0.6} & \meanpmstddevbold{90.1}{0.3} &\meanpmstddevbold{50.2}{0.5}\\
        \addlinespace
        ProPos & 94.3 & 53.2 & 86.7 &95.6 &74.5\\
        DCvCD (ProPos) & \meanpmstddevbold{95.3}{0.2} & \meanpmstddevbold{54.9}{0.9} & \meanpmstddevbold{88.9}{0.3} & \meanpmstddevbold{96.0}{0.1}&\meanpmstddevbold{75.8}{0.3} \\
        \addlinespace
        CoNR & 93.2 & 53.6 & 92.6 & 96.4 &79.4 \\
        DCvCD (CoNR) & \meanpmstddevbold{94.2}{0.3} & \meanpmstddevbold{55.8}{0.6} & \meanpmstddevbold{94.0}{0.2} & \meanpmstddevbold{97.5}{0.1} &\meanpmstddevbold{81.0}{0.1}\\ 
        \addlinespace
        TCL & 88.7 & 51.2 & 86.8 & 89.5 &64.4 \\
        DCvCD (TCL) & \meanpmstddevbold{90.9}{1.1} & \meanpmstddevbold{53.7}{1.5} & \meanpmstddevbold{88.6}{0.7} & \meanpmstddevbold{92.5}{0.6} &\meanpmstddevbold{66.2}{0.9}\\
        \bottomrule
    \end{tabular}
\end{table*}

\begin{table*}[htbp]
    \centering
    \small
        \caption{The ablation evaluation results of the community merging metrics.}
    \label{tab4}
    \begin{tabular}{lccccccccccccccc}
        \toprule
        \multirow{2}{*}{Method} 
            & \multicolumn{3}{c}{CIFAR-10(\%)} 
            & \multicolumn{3}{c}{CIFAR-100(\%)} 
            & \multicolumn{3}{c}{STL-10(\%)} 
            & \multicolumn{3}{c}{ImageNet-10(\%)} 
            & \multicolumn{3}{c}{ImageNet-Dogs(\%)} \\
        \cmidrule(lr){2-4} \cmidrule(lr){5-7} \cmidrule(lr){8-10}
        \cmidrule(lr){11-13} \cmidrule(lr){14-16}
             & NMI & ACC & ARI 
            & NMI & ACC & ARI 
            & NMI & ACC & ARI 
            & NMI & ACC & ARI 
            & NMI & ACC & ARI \\
        \midrule
        DCvCD (w/o $\Delta Q$) 
            & 85.5 & 91.5 & 84.0  
            & 50.1 & 52.1 & 35.0 
            & 84.0 & 90.8 & 82.1 
            & 88.8 & 93.6 & 88.4 
            & 72.5 & 77.0 & 63.7 \\
        DCvCD (w/o $\Delta k$) 
            & 87.1 & 92.8 & 86.1 
            & 52.6 & 54.7 & 37.8 
            & 86.2 & 92.9 & 84.5 
            & 91.0 & 95.9 & 91.5 
            & 75.2 & 79.9 & 65.9 \\
        DCvCD (w/o $t$)  
            & 86.5 & 92.4 & 85.3 
            & 51.9 & 53.8 & 36.9 
            & 85.5 & 92.1 & 83.7 
            & 90.5 & 95.5 & 91.0 
            & 75.0 & 79.3 & 65.1 \\
        DCvCD (Full) 
            & \textbf{88.7} & \textbf{94.2} & \textbf{87.8}  
            & \textbf{53.6} & \textbf{55.8} & \textbf{39.0}   
            & \textbf{87.2} & \textbf{94.0} & \textbf{85.8}  
            & \textbf{92.0} & \textbf{97.5} & \textbf{93.1}  
            & \textbf{76.2} & \textbf{81.0} & \textbf{67.4}  \\
        \bottomrule
    \end{tabular}
\end{table*}

\begin{table}[h!]
    \centering
    \small
        \caption{Runtime comparison between DCvCD and ProPos: averages and variances over 10 runs.}
    \label{tab:runtime}
    \resizebox{\columnwidth}{!}{%
    \begin{tabular}{@{}ccccc@{}}
        \toprule
        Method & CIFAR-10(h) & CIFAR-100(h) & STL-10(h) & ImageNet-10(h)\\
        \midrule
        DCvCD  & $9.3 \pm 0.9$ & $10.2 \pm 1.7$ & $3.1 \pm 0.7$ & $4.0 \pm 1.1$  \\
        ProPos & $10.2 \pm 1.7$ & $10.4 \pm 1.8$ & $14.2 \pm 0.9$ & $2.5 \pm 1.4$ \\
        \bottomrule
    \end{tabular}
    } 
\end{table}

It is noteworthy that for fair comparison, we do not directly compare DCvCD with the existing text-enhanced approaches, e.g., CLIP~\cite{radford2021learning}, SIC~\cite{cai2023semantic} and TAC~\cite{li2023image}, since they leverage additional external knowledge to facilitate clustering. However, we also present their performance results for reference purposes in the appendix due to space limit. The experimental results show that the performance of DCvCD is competitive compared with these text-enhanced approaches.





 

  In the comparative study, our implementations of DCvCD employ DivClust, which is a primitive but popular algorithm, and ProPos and CoNR, which have reported SOTA results on the test datasets, for initial clustering, and their RL backbones for representation fine-tuning. ProPos and CoNR use the ResNet-34 and the ResNet-18 as their RL backbones respectively. For model fine-tuning, we use the SGD optimizer with an initial learning rate of 0.0001 to balance training stability and convergence speed. We set the batch size at 64, and the number of epochs at 100. In community detection and merging, the threshold of representation similarity for edge addition is set to the default value of 0.5. Our experimental results, which are put in the appendix due to space limit, show that the performance of DCvCD is robust w.r.t. the similarity threshold provided that it is set within a reasonable range. 
  We have conducted all the experiments on an NVIDIA RTX A6000 GPU. As in previous work, we report the performance results averaged over 10 runs in the experiment.

\subsection{Comparative Evaluation}

The detailed comparative results have been reported in Table~\ref{tab1}, in which the best results are \textbf{highlighted}. It can be observed that our proposed approach of DCvCD improves the SOTA performance on all the five datasets. Among the existing alternatives without leveraging external knowledge, ProPos and CoNR achieve the overall best performance. The corresponding DCvCD solutions still manage to improve performance. This observation clearly demonstrates that the proposed clustering strategy is complementary to the existing SOTA approaches. 

Notably, compared to the more primitive approach of DivClust, the corresponding DCvCD(DivClust) solution shows considerable improvement across all the metrics. For instance, in terms of ACC, DCvCD(DivClust) outperforms DivClust by more than 8\% on CIFAR-10, CIFAR-100 and STL-10. 

It is interesting to point out that the improvement margins of DCvCD are not uniform across datasets: they tend to be larger when the initial clustering is relatively weak and more modest when the baseline is already very strong. Overall, the evaluation results clearly demonstrate that DCvCD, which introduces the
perspective of network analysis into the clustering strategy, can effectively improve image clustering performance.


\begin{table*}[h!]
    \centering
    \small
    \caption{Performance comparison of different community detection algorithms implemented in DCvCD(CoNR).}
    \label{tab:community_algo_compare}
    \begin{tabular}{lccccccccccccccc}
        \toprule
        \multirow{2}{*}{Algorithm} 
            & \multicolumn{3}{c}{CIFAR-10(\%)} 
            & \multicolumn{3}{c}{CIFAR-100(\%)} 
            & \multicolumn{3}{c}{STL-10(\%)} 
            & \multicolumn{3}{c}{ImageNet-10(\%)} 
            & \multicolumn{3}{c}{ImageNet-Dogs(\%)} \\
        \cmidrule(lr){2-4} \cmidrule(lr){5-7} \cmidrule(lr){8-10} \cmidrule(lr){11-13} \cmidrule(lr){14-16}
            & NMI & ACC & ARI 
            & NMI & ACC & ARI 
            & NMI & ACC & ARI 
            & NMI & ACC & ARI 
            & NMI & ACC & ARI \\
        \midrule
        CoNR (Baseline)    
            & 86.7  & 93.2  & 86.1  
            & 51.3  & 53.6  & 37.5   
            & 85.2  & 92.6  & 84.6  
            & 91.1  & 96.4  & 92.2    
            & 74.4  & 79.4  & 66.7    \\
        \midrule
        Louvain            
            & 87.0  & 93.6  & 86.5  
            & 51.9  & 54.5  & 37.9 
            & 85.7  & 93.4  & 85.0  
            & 91.3  & 96.9  & 92.5  
            & 74.8  & 80.2  & 67.0 
            \\
        \midrule
        GN                 
            & 86.8  & 93.4  & 86.2  
            & 51.5  & 54.0  & 37.6  
            & 85.3  & 92.9  & 84.7 
            & 91.2  & 96.6  & 92.3 
            & 74.5  & 79.8  & 66.8  
            \\
        \midrule
        DBSCAN             
            & 88.0  & 93.9  & 87.0  
            & 52.7  & 55.1  & 38.2 
            & 86.9  & 93.6  & 85.5 
            & 91.8  & 97.2  & 92.8  
            & 75.5  & 80.6  & 67.3  
            \\
        \midrule
        Leiden (Ours)      
            & \textbf{88.7}  & \textbf{94.2} & \textbf{87.8}  
            & \textbf{53.6}  & \textbf{55.8} & \textbf{39.0}  
            & \textbf{87.2}  & \textbf{94.0} & \textbf{85.8}  
            & \textbf{92.0}  & \textbf{97.5} & \textbf{93.1} 
            & \textbf{76.2}  & \textbf{81.0} & \textbf{67.4} 
            \\
        \bottomrule
    \end{tabular}
\end{table*}

\subsection{Ablation Study}




 \noindent\textbf{Clustering Strategy:} to verify the benefit of the community-oriented clustering strategy, we implement the DCvCD solution based on various representative clustering methods, including CC (Contrastive Clustering)~\cite{li2021contrastive} and TCL (Twin Contrastive Learning)~\cite{li2022twin}, besides DivClust, ProPos and CoNR as reported in Table \ref{tab1}, and compare their performance with its corresponding baselines in terms of ACC. To verify the statistical significance of DCvCD's superior performance, we also report its performance variances besides the averages over 10 runs. W.r.t. the compared baselines, since only the ProPos reports performance variances, we report both averages and variances on ProPos but only averages on other baselines. 
 
The detailed comparison and evaluation results are presented in Table~\ref{tab2}. It can be observed that DCvCD consistently outperforms its corresponding baseline on all the datasets. It is noteworthy that its margins in terms of averages comfortably exceed variances, validating statistical significance of DCvCD advantage. Furthermore, it is interesting to point out that the improvement margins are overall more considerable if the baseline underperforms. On three of the five datasets, CIFAR-10, CIFAR-100 and ImageNet-10, even beginning with the suboptimal clustering results of DivClust, DCvCD(DivClust) achieves competitive performance compared with DCvCD(ProPos) and DCvCD(CoNR). Our experimental results clearly demonstrate that the proposed community-oriented clustering strategy can effectively improve the performance of existing RL approaches. They bode well for its application in real scenarios.

\noindent\textbf{Community Merging Metrics:} to verify the efficacy of the three metrics used in community merging, we conduct an ablation study on their impact on the performance of DCvCD. The detailed evaluation results on DCvCD(CoNR) have been presented in Table \ref{tab4}. The results on other DCvCD solutions are similar, thus omitted here. It can be observed that removing any of the three metrics, i.e., the modularity, the average degree and the distance, the performance of DCvCD would drop, with the modularity having the biggest adverse impact. These observations clearly demonstrate that the three metrics are complementary to each other and their ensemble achieves the best performance.

\subsection{Efficiency Evaluation}





To evaluate the efficiency of DCvCD clustering strategy, we compare its overhead with ProPos, one of only a few SOTA works that report runtime results. The detailed comparison is presented in Table~\ref{tab:runtime}, in which the overhead refers to the runtime of DCvCD after initial clustering is completed. The results are averages and variances over 10 runs.

It can be observed that: 1) on CIFAR-10, CIFAR-100, and ImageNet-10, the runtime of DCvCD is comparable to that of ProPos; 2) on the STL-10 dataset, DCvCD is significantly more efficient than ProPos. The main reason is that ProPos uses an additional 100,000 unlabeled images to optimize representation learning, whereas community detection and merging does not involve these extra data. These results demonstrate that the overhead of DCvCD is comparable to that of the existing clustering algorithms. 




\subsection{Choices of Community Detection Algorithm}

To verify the efficacy of DCvCD w.r.t. different community detection algorithms and the superiority of the Leiden algorithm, we have conducted a comparative study using four representative algorithms: Leiden (the default), Louvain \cite{blondel2008fast} (the traditional modularity-based algorithm), GN \cite{girvan2002community} (a graph-splitting-based algorithm), and DBSCAN \cite{ester1996density} (a density-based algorithm). All the implementations are based on the CoNR, since CoNR achieves the best baseline performance. 

We have presented the detailed evaluation results in Table \ref{tab:community_algo_compare}. It can be observed that:
\begin{itemize}
\item DCvCD can effectively improve performance using any tested community detection algorithm. This result clearly validates the value of gradual community detection to cluster optimization, demonstrating the generality and flexibility of our proposed approach.

\item Our choice of Leiden algorithm consistently achieves the best performance on clustering quality.  Leiden's superior performance can be attributed to its explicit refinement of partitions and enforcement of well-connected communities, which can effectively reduce poorly connected or fragmented clusters. Consequently, the communities found by Leiden can usually better align with semantic classes, and thus provide a stronger starting point for gradual community merging.

\end{itemize}





\section{Conclusion and Future Work}


In this paper, we propose a novel approach of deep clustering via gradual community detection, whose clustering strategy factors in cluster network analysis. Our extensive experiments on benchmark image datasets have validated its efficacy. It is noteworthy that the proposed DCvCD, as a clustering strategy, is a flexible approach. It can be easily built upon the existing representation learning approaches for deep clustering. The flexibility of DCvCD bodes well for its application in real scenarios.

On future work, it is worthy to extend DCvCD beyond the image-only setting by applying gradual community detection on multimodal similarity graphs, so that unsupervised clustering can benefit from complementary visual and textual information. Our current work separates the stages of representation learning and community detection. This strategy has the obvious benefit that it can easily leverage the existing RL backbones. However, in future work, it is interesting to investigate how to effectively align RL backbone design and training with community characteristics.

\small
\bibliographystyle{unsrt}  
\bibliography{references}  


\end{document}